\pdfoutput=1

\documentclass[11pt]{article}

\usepackage{EACL2023}

\usepackage{times}
\usepackage{latexsym}

\usepackage[T1]{fontenc}

\usepackage[utf8]{inputenc}

\usepackage{microtype}

\usepackage{inconsolata}

\usepackage{tabularx}
\usepackage{graphicx}
\usepackage{multirow}
\usepackage{booktabs}
\usepackage{enumitem}
\usepackage{multirow}
\usepackage{xspace}
\newcommand{\F}{F$_1$\xspace}
\usepackage{caption}
\usepackage{amsmath}
\usepackage{hyperref}
\usepackage{graphicx}
\usepackage{microtype}
\newcommand{\rt}[1]{\rotatebox[origin=c]{90}{#1}}
\makeatletter
\renewcommand{\paragraph}{%
  \@startsection{paragraph}{4}%
  {\z@}{0.25ex \@plus 1ex \@minus .2ex}{-1em}%
  {\normalfont\normalsize\bfseries}%
}
\makeatother

\title{An Entity-based Claim Extraction Pipeline for\\ Real-world Biomedical Fact-checking}

\author{Amelie W\"uhrl, Lara Grimminger, \and Roman Klinger \\
  Institut f{\"u}r Maschinelle Sprachverarbeitung, University of Stuttgart, Germany \\
  \texttt{\{amelie.wuehrl, lara.grimminger,}\\\texttt{roman.klinger\}@ims.uni-stuttgart.de}\\
}

\begin{document}
\maketitle
\begin{abstract}
Existing fact-checking models for biomedical claims are typically trained on synthetic or
  well-worded data and hardly transfer to social media content. This
  mismatch can be mitigated by adapting the social media input to
  mimic the focused nature of common training claims. To do so,
  \newcite{wuehrl-klinger-2022} propose to extract concise claims
  based on medical entities in the text. However, their study has two
  limitations: First, it relies on gold-annotated entities.
  Therefore, its feasibility for a real-world application cannot be
  assessed since this requires detecting relevant entities
  automatically. Second, they represent claim entities with the
  original tokens. This constitutes a terminology mismatch which
  potentially limits the fact-checking performance. To understand both
  challenges, we propose a claim extraction pipeline for medical
  tweets that incorporates named entity recognition and terminology
  normalization via entity linking. We show that automatic NER does
  lead to a performance drop in comparison to using gold annotations
  but the fact-checking performance still improves considerably over
  inputting the unchanged tweets. Normalizing entities to their
  canonical forms does, however, not improve the performance.
\end{abstract}

\section{Introduction}
\begin{figure*}[tb]
\centering
\includegraphics[scale=.8]{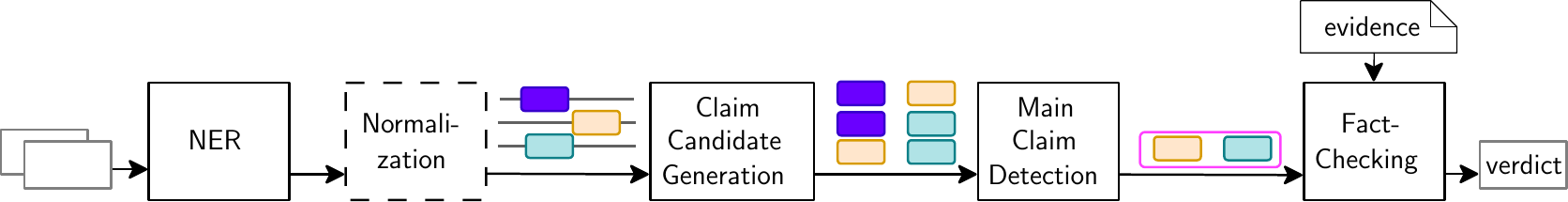} 
\caption{Overview of the claim extraction pipeline. Input documents go
  through entity recognition (NER), normalization, claim candidate
  generation, main claim detection and fact-checking. Colored boxes represent the entities which we use to
  extract claim candidates. Note that we evaluate the normalization
  module separately from the evaluation of the rest of the pipeline
  (see \S\ref{sec:experiments}).}
\label{fig:pipeline}
\end{figure*}
Fact-checking models trained on synthetic, well-worded and atomic
claims struggle to transfer to colloquial content
\citep{kim-etal-2021}. There are multiple ways to address this
problem: We can build custom datasets and models that verify medical
content shared online
\citep{saakyan-etal-2021,mohr-etal-2022,sarrouti-etal-2021} and tackle
related tasks \citep{sundriyal-et-al_2022,dougrez-lewis-etal-2022}.
Alternatively, we can adapt the input before addressing other
fact-checking tasks. \newcite{bhatnagar-etal-2022} create claim
summaries and find that this improves the detection of previously
fact-checked claims. Similarly, \newcite{wuehrl-klinger-2022} extract
concise claims from user-generated text in an effort to mimic the
focused, well-structured nature of the claims the fact-checking models
were originally trained on. They find that this improves the accuracy
of pre-trained evidence-based fact-checking models in the biomedical domain.

\begin{table}
\small
\begin{tabularx}{\columnwidth}{lXp{33mm}}
  \toprule
  & Claim & Evidence \\
  \cmidrule(r){1-2}\cmidrule(l){3-3}
  \multirow{2}{*}{\rt{orig}}
  &
    medicines causes blood clots
          &
            \multirow{2}{*}{\parbox{32mm}{drospirenone may
            significantly increase
            chances of developing
            venous thromboembolic
            events}}\\
  \cmidrule(r){1-2}
  \multirow{3}{*}{\rt{norm}} & pharmaceutical preparations causes thrombus\\
  \bottomrule
\end{tabularx}
\caption{Example claim represented with original and normalized
  entities together with evidence.}
\label{table:normalization-example}
\end{table}
However, the study by \newcite{wuehrl-klinger-2022} is limited in two ways:
(1) Their claim extraction method relies on gold-annotated,
claim-related entities. For a realistic evaluation, such an oracle needs
to be replaced by an entity recognizer. Only then it is possible measure the impact of
potential error propagation which may ultimately render the method
unfeasible.
(2) The claim entities are represented by the original token
sequence. This is problematic as medical mentions on Twitter
potentially contain imprecise, abbreviated, or colloquial
terminology. This is in contrast to the terminology in the original
model input as well as the documents that we provide as evidence (cf.\
Table~\ref{table:normalization-example}). We hypothesize that for a
successful fact-check we need to close this gap by normalizing medical
terminology in the input. Previous work suggested leveraging entity
linking for evidence retrieval
\citep{nooralahzadeh-ovrelid-2018,taniguchi-etal-2018,hanselowski-etal-2018}
leading us to believe that it could also be beneficial for aligning
claim and evidence.

We address both limitations and evaluate a real-world, fully-automatic
claim extraction pipeline for medical tweets which incorporates an
entity recognizer. It only relies on the original text as input that
contains the claim. We further evaluate the impact of an entity linker
for normalizing entity mentions to canonical forms based on the
Unified Medical Language System \citep[UMLS,][]{bodenreider-2004}.
Our pipeline improves the fact-checking performance over tasking
models to check unchanged tweets.  Normalizing entities to overcome
the terminology mismatch does not improve fact-checking,
potentially due to limitations of biomedical entity linking for social
media.

\section{Methods}
\label{sec:methods}
Figure~\ref{fig:pipeline} visualizes our pipeline. It takes text as
input and performs \textit{named entity recognition} and optionally
term \textit{normalization} via entity linking. Each unique entity
pair forms the building blocks for a potential claim (\textit{claim
  candidate generation}). The \textit{main claim detection} identifies
the core claim among the candidates that presumably represents the
most important aspect of the text. The resulting claim is the input to
the fact-checker. In our setting, we assume this to be a frozen
pre-trained fact-checking model. We describe the modules in the
following and the fact-checker in Section~\ref{sec:evaluation}.

\paragraph{NER.} We use the SpaCy
environment\footnote{\url{https://spacy.io/api/architectures\#TransitionBasedParser}}
to train a custom NER model that detects medical entities. This
framework relies on a transition-based parser \citep{lample-etal-2016}
to predict entities in the input.  In a preliminary study, we found
that relying on an off-the-shelf model for biomedical NER, i.e.,
ScispaCy \citep{neumann-et-al-2019}, does not transfer to medical
texts from social media. Refer to Appendix
\ref{appendix:analysis-eval-setup} for a comparison of the two models.

\paragraph{Claim candidate generation.}
\newcite{wuehrl-klinger-2022} propose two extraction methods, i.e.,
$\textrm{condense}_{\textrm{seq}}$ and
$\textrm{condense}_{\textrm{triple}}$. The first represents the claim
as the token sequence from the first entity to the last entity, while
the second relies on gold-annotated causal relations which they use to
build the claims. We use the sequence method
$\textrm{condense}_{\textrm{seq}}$ in our pipeline because
both methods show on par performances (difference in 1pp
\F) and, in contrast to $\textrm{condense}_{\textrm{triple}}$, it
does not require relation classification.

Following the $\textrm{condense}_{\textrm{seq}}$ method, we therefore
extract the sequence from the character onset of the first entity to the character offset
of the second entity for all pairs of entities found by the NER
module.

\paragraph{Entity linking.}
To normalize entities, we use the \textit{EntityLinking} component in
ScispaCy \citep{neumann-et-al-2019}. This model compares an entity
mention to concepts in an ontology and creates a ranked list of
candidates, based on an approximate nearest neighbor search. For text
normalization, we retrieve the canonical name of the top concept. For
entities which could not be linked, we use the original mention
instead. As the knowledge base, we use UMLS \citep{bodenreider-2004}.

\paragraph{Main claim detection.}
\label{sec:main-claim}
For tweets with more than two predicted entities, claim generation
produces multiple claim candidates. To identify the claim to 
be passed to the fact-checking module, we train a text classifier to
detect the main claim for a given input. We build on
\textsc{R}o\textsc{BERTA}rg\footnote{\url{https://huggingface.co/chkla/roberta-argument}},
a RoBERTA-based text classification model trained to label input texts
as \textsc{argument} or \textsc{non-argument}. We fine-tune this model
to classify texts as \textsc{claim} vs.\ \textsc{non-claim} and to fit
the social media health domain. At inference time, the claim candidate with
the highest probability for the claim class constitutes the main
claim. We refer to this as \textit{ner$+$core-claim}.

\begin{table*}[]
\small\centering
    \setlength{\tabcolsep}{5pt} 
    \begin{tabularx}{\textwidth}{lrrrrrrrrrrrrrrr}
    \toprule
        & \multicolumn{15}{c}{Input Claim}\\
        \cmidrule(l){2-16}
        & \multicolumn{4}{c}{Gold entities}&&&&\multicolumn{8}{c}{Fully automatic (Ours)}\\
        \cmidrule(lr){2-5} \cmidrule(l){9-16}

        & \multicolumn{4}{c}{$\textrm{condense}_{\textrm{seq}}$} 
        & \multicolumn{3}{c}{full tweets} 
		&  \multicolumn{4}{c}{ner$+$rand-ent-seq}       
		&  \multicolumn{4}{c}{ner$+$core-claim}        
        \\
        \cmidrule(lr){2-5} \cmidrule(lr){6-8} \cmidrule(lr){9-12} \cmidrule(lr){13-16} 
        model& P & R & \F & $\Delta_\textrm{full}$ & P & R & \F & P & R & \F & $\Delta_\textrm{full}$  & P & R & \F & $\Delta_\textrm{full}$  \\
         
         \cmidrule(r){1-1}
      \cmidrule(lr){2-2}\cmidrule(lr){3-3}\cmidrule(lr){4-4}
      \cmidrule(lr){5-5}\cmidrule(lr){6-6}\cmidrule(lr){7-7}\cmidrule(lr){8-8}
      \cmidrule(lr){9-9}\cmidrule(lr){10-10}\cmidrule(lr){11-11}\cmidrule(lr){12-12}
      \cmidrule(lr){13-13}\cmidrule(lr){14-14}\cmidrule(lr){15-15}\cmidrule(l){16-16}

    		fever & 
    		83.3 & 1.9 & 3.7 & $+$3.7 
    		& 0.0 & 0.0 & 0.0 
    		& 0.0 & 0.0 & 0.0  & $+$0
    		& 100 & 0.4 & 0.8 & $+$0.8
    		\\
    		
    		fever\_sci & 
    		87.2 & 15.5 & 26.4 & $+$18.4 
    		& 91.7 & 4.2 & 8.0 
    		& 92.3 & 4.7 & 9.0 & $+$1.0 
    		& 82.4 & 5.6 & 10.4 & $+$2.4
    		\\

    		scifact & 
    		90.9 & 7.6 & 14.0 & $+$13.2 
    		& 100 & 0.4 & 0.8 
    		& 100 & 2.4 & 4.6 & $+$3.8
    		& 100 & 2.4 & 4.7 & $+$3.9
    		\\
    		
    		covidfact & 
    		55.6 & 28.4 & 37.6 & $+$29.7
    		& 30.8 & 4.5 & 7.9 
    		& 53.3 & 9.4 & 16.1 & $+$8.2
    		& 58.1 & 14.3 & 23.0 & $+$15.1
    		\\

    		healthver & 
    		85.9 & 48.5 & 62.0 & $+$16.8 
    		& 82.8 & 31.1 & 45.2
    		& 75.6 & 23.2 & 35.5 & $-$9.7    		
    		& 77.4 & 28.7 & 41.9 & $-$3.3
    		\\
    		
    		\cmidrule(lr){2-5} \cmidrule(lr){6-8} \cmidrule(lr){9-12} \cmidrule(lr){13-16} 
		average & 80.6 & 20.4 & 28.7 & $+$16.3	
    		& 61.1 & 8.0 & 12.4 
    		& 64.2 & 7.9 & 13.0& $+$0.6
		& 83.6 & 10.1& 16.2& $+$3.8
    		\\

    \bottomrule
    \end{tabularx}
    \caption{Performance (precision, recall
      and \F) of MultiVerS-based models
      (\textit{fever, fever\_sci, scifact, covidfact, healthver}) on
      Co\textsc{Vert} data. Model inputs are the
      full tweets, the entity-based sequence claims ($\textrm{condense}_{\textrm{seq}}$  \citep{wuehrl-klinger-2022}), and claims from the fully automatic pipeline, \textit{ner$+$rand-ent-seq} and \textit{ner$+$core-claim}. $\Delta_\textrm{full}$ : difference in \F between the full tweet and performance for the respective input claim. We report the average across all models in the last row.}
    \label{tab:multivers-on-covert}
\end{table*}

\section{Experiments}
\label{sec:experiments}
\subsection{Data}
\label{sec:data}

\paragraph{Co\textsc{Vert}.} We use the Co\textsc{Vert} dataset
\citep{mohr-etal-2022} to test our pipeline. It consists of medical
tweets labeled with fact-checking verdicts (\textsc{Supports},
\textsc{Refutes}, \textsc{not enough information}) and associated
evidence texts. We follow the same filtering and preprocessing as
\newcite{wuehrl-klinger-2022} which leaves us with 264 tweets. For 13
tweets, the NER model predicts only one or no entities. In these
cases, we cannot generate claim candidates thus we can only consider 251
claims.

\paragraph{\textsc{Bear}.} We require an independent dataset to train
the NER component. We find the \textsc{Bear} dataset
\citep{wuhrl-klinger-2022a} to be closest in domain and text type to
the target data from Co\textsc{Vert}. \textsc{Bear} provides 2100
tweets with a total of 6324 annotated medical entities from 14
entity classes. We use 80\% of the data for training and 20\% for
testing the model.

\paragraph{Causal Claims.} To build a classifier that identifies the
core claims, we use the \textsc{Causal Claims} data from SemEval-2023
Task 8, Subtask 1.\footnote{\url{https://causalclaims.github.io/}} It
consists of medical Reddit posts and provides span-level annotations
for \textit{Claim}, \textit{Experience}, \textit{Experience based
  claim} and \textit{Question}. Our goal is to differentiate claims
from non-claims. Consequently, we extract all spans labeled as
\textit{Claim} and \textit{Experience based claim} as positive
instances for the claim class and use the remaining text spans as
negative examples. This leads to 1704 claim and 6870 non-claim
spans. We use a train/test split of 90/10\%.

\subsection{Evaluation}
\label{sec:evaluation}
The fact-checking module serves as a by-proxy evaluation for the claim
representations.  Provided with a claim--evidence pair, the system
predicts a fact-checking verdict that indicates if the evidence
\textsc{Supports} or \textsc{Refutes} the claim.  We assume that the
fact-checker is a frozen model for which we adapt the claim input.  To
gauge the checkability of a particular input, we compare the
performance for predicting the correct verdict when the model is
presented with claims of this type. This follows the evaluation in
\newcite{wuehrl-klinger-2022}.

The fact-checking models we employ stem from the MultiVerS
architecture
\citep{wadden-etal-2022}.\footnote{\url{https://github.com/dwadden/multivers}}
This framework is designed for scientific fact-verification and
provides five models (\textit{fever, fever\_sci, scifact, covidfact,
  healthver}), differing in training data.  We report precision,
recall and \F for predicting the correct fact-checking verdict
(\textsc{Supports}, \textsc{Refutes}, \textsc{not enough information})
for a given claim-evidence pair.

\subsection{Exp.~1: Impact of NER}
In Exp.~1, we aim to understand the impact of automatic NER and
main claim detection in the pipeline, instead of relying on
gold-labeled entities.

Table \ref{tab:multivers-on-covert} reports the results for our fully
automatic claim extraction pipeline.  Each column reports the
performance for a specific type of input claim. \textit{Full tweets}
is the performance as reported by \newcite{wuehrl-klinger-2022} for
the unchanged input tweets. The results denoted with
$\textrm{condense}_{\textrm{seq}}$ describe their results with gold
annotations, to which we compare. Our main results are in the last
column (\textit{ner$+$core-claim}). To
understand the impact of the main claim detection, we compare against a purely
random selection of the main claim from all candidates in the tweet (\textit{ner$+$rand-ent-seq}).

The rows correspond to the various fact-checking models.
 $\Delta$ columns report the difference in \F between the
performance of checking the full tweet and the respective claim
representation.

\textit{ner$+$core-claim} shows an average performance of \F=16.2. The
performance varies across the models. The \textit{healthVer} model
performs the best (41.9\F). The average is considerably higher than
using the full tweets ($\Delta$=3.8\,pp\,\F). This improvement is
consistent across all models, except for \textit{healthVer},
presumably because it already shows a high performance for the
original texts. To better understand the model behavior, we provide an analysis of its prediction in Appendix~\ref{sec-error-analysis-healthver}. We see a particularly strong impact for the
\textit{covidfact} model, with $\Delta$=15.1\,pp. Despite this
positive result, we see a performance drop when integrating entity
recognition instead of building claim extraction on gold entity
annotations. This decrease is not surprising since we expect some
error propagation from an imperfect entity recognizer. Nevertheless,
the results show that entity-based claim extraction also increases the
fact-checking performance even under some error propagation throughout
the real-world pipeline.

We further see that main claim detection is a required module -- the
performance for a randomly selected claim
(\textit{ner$+$rand-ent-seq}) is substantially lower. This indicates
that using the same evidence and fact-checking model, not all
potential claims in a tweet would receive the same verdict.

\subsection{Exp.~2: Impact of Entity Normalization}
In Exp.~2, we investigate if it is beneficial to
assimilate the linguistic realizations of medical mentions to the
expected input of the fact-checking models. More specifically, we
suggest normalizing entity strings in the input. In contrast to
Exp.~1, in which we evaluate the overall pipeline, we focus on the
aspect of the entities here and therefore do not make use of the core
claim detection method or the entity recognizer. Instead we build on
top of gold annotations and, consequently, employ
$\textrm{condense}_{\textrm{triple}}$ described in Section~\ref{sec:methods}.

We use entity linking for term normalization and use ScispaCy's entity
linking functionality with \textit{en\_core\_sci\_sm} as the
underlying model \citep{neumann-et-al-2019}.  For each (gold) entity,
we use the canonical name of the concept with the highest linking
score. Subsequently, we follow the
$\textrm{condense}_{\textrm{triple}}$ method to represent claims.

Table \ref{tab:normalization} reports the results for claims built
with non-normalized (\textit{surface string}) vs.\ normalized entities
(\textit{normalized ent.}). The results indicated as
$\textrm{condense}_{\textrm{triple}}$ \textit{surface string} are
analogue to the results in \newcite{wuehrl-klinger-2022}. We see that
normalization does not have the desired effect: The verdict prediction
performance drops across all of the fact-checking models (from 29.7 to
22.6 in avg.~\F). We assume that this is, to a considerable extend,
due to entity linking being a challenging task which leads to a
limited performance of the employed linking module. We present an
error analysis in Appendix~\ref{sec:analysis-linking}.

\begin{table}[]
\small \centering
    \setlength{\tabcolsep}{5pt} 
    \begin{tabular}{lrrrrrr}
    \toprule
        & \multicolumn{6}{c}{$\textrm{condense}_{\textrm{triple}}$ Claims}\\
        
        \cmidrule(lr){2-7}  
        & \multicolumn{3}{c}{surface string} 
        & \multicolumn{3}{c}{normalized ent.} 
      
        \\
        \cmidrule(lr){2-4} \cmidrule(lr){5-7}
        model& P & R & \F & P & R & \F \\
         
         \cmidrule(r){1-1}
      \cmidrule(lr){2-2}\cmidrule(lr){3-3}\cmidrule(lr){4-4}
      \cmidrule(lr){5-5}\cmidrule(lr){6-6}\cmidrule(lr){7-7}

    		fever & 
    		81.8 & 3.4 & 6.5
    		& 75.0 & 1.1 & 2.2
    		\\
    		
    		fever\_sci & 
    		89.8 & 20.1 & 32.8
    		& 93.9 & 11.7 & 20.9
    		\\

    		scifact & 
		86.4 & 7.2 & 13.3
		& 94.4 & 6.4 & 12.1
    		
    		\\
    		
    		covidfact & 
		65.0 & 30.3 & 41.3
		& 61.8 & 20.8 & 31.2
 
    		\\

    		healthver & 
		79.7 & 41.7 & 54.7  
		& 85.7 & 31.8 & 46.4
		
    		\\
    		
    		\cmidrule(lr){2-4} \cmidrule(lr){5-7} 
    		average 
    		& 80.5 & 20.5 & 29.7
		& 82.2& 14.4  &22.6
    		\\

    \bottomrule
    \end{tabular}
    \caption{Performance (precision, recall
      and \F) of MultiVerS-based fact-checking models
      (\textit{fever, fever\_sci, scifact, covidfact, healthver}) on
      Co\textsc{Vert} claims built with non-normalized (surface string) vs. normalized entities. We report the average across all models in the last row.}
    \label{tab:normalization}
\end{table}

\section{Conclusion \& Future Work}
We propose a fully automatic claim extraction pipeline that is capable
of handling real-world medical content. We show that entity-based
claim extraction has a positive effect on the performance of multiple
fact-checking models -- even after replacing the entity oracle with
automatic NER. While we observe a negative impact of error propagation
from NER and a performance drop as a result, fact-checking the
extracted claims is more successful than checking
unchanged tweets. Future research may therefore
focus on improving the pipeline components as this clearly has the
potential to further strengthen the verdict prediction performance. In particular, we
expect an improved entity recognizer to have a
considerable impact.

Our work focuses on the biomedical domain and builds upon the
assumption by \newcite{wuehrl-klinger-2022} that claims in this domain
are strongly centered around entities. Claims from other domains may
share this property which could make entity-based claim extraction
applicable for such claims as well. We leave the evaluation for future
work.

We find that normalizing entity mentions does not improve the
fact-checking performance. However, our analysis shows that the
off-the-shelf linking module might be too unreliable. To fully gauge the
potential of normalizing entities, future work needs to ensure correct
mappings (creating gold links or building a reliable linker) before
evaluating the downstream fact-checking performance.

\section*{Acknowledgments}
This research has been conducted as part of the FIBISS project which is funded by the German Research Council (DFG, project number: KL~2869/5-1). We thank the anonymous reviewers for their valuable feedback.

\section*{Limitations}
Our work focused on evaluating the impact of putting together a set of
components to achieve a real-world system for fact-checking. For
answering the research question at hand, the components offered
themselves as appropriate choices. This being said, to some degree, the particular selection may limit the expressiveness of the experiments.

By instantiating the pipeline components with the set of models and
underlying data that we chose, our findings are limited to this
setting. However, the analysis that we provide in Appendix
\ref{appendix:analysis} dissects the pipeline results and allows us to
draw more general conclusions about the impact of replacing individual
components.

We propose that the main claim detection receives more attention in
future research. This may mitigate the issue that this module is potentially
the most in-transparent component. Compared to the NER, this task can
be modeled in various ways. We rely on the output probabilities to
identify the claim candidate the model is most confident about. While
this is a straight-forward approach and we show that it works as
intended, prediction probabilities -- especially for deep models --
may not always be a distinctive indicator of model confidence. To
overcome this limitation, alternative ways of detecting the main claim
should be evaluated.

\section*{Ethical Considerations}
A real-world fact-checking pipeline presents itself as a valuable
tool. However, we advise against using the pipeline purely automatically that at this point in time. Unless they are used hand-in-hand with a human expert performing or supervising the fact-check, such systems are not reliable enough yet.

Potential issues are the result of the inherent opaqueness of sophisticated
automatic analysis pipelines. In the system that we propose, it is
important that the impact of each module needs to explain itself to
the user. While there is recent work on explainability particularly in
the area of fact checking, this work did not yet focus on entity-based approaches. It is important that a user can clearly
understand which claim in a statement is checked and which risks
potential error propagation might lead to. Therefore, before deploying
such systems for fully automatic filtering or labeling of problematic
messages in a social media content, there needs to be more research on
explainability and transparency of such systems.

\bibliographystyle{acl_natbib}
\bibliography{literature}

\appendix

\section{Implementation details}
\label{appendix-training-details}
In the following, we provide implementation details for the individual model components described in Section \ref{sec:methods}.
\subsection{Named Entity Recognition}
In a preliminary experiment, we use a pre-trained model for biomedical NER, i.e., the \textit{en\_core\_sci\_sm} model by ScispaCy \citep{neumann-et-al-2019}, that was trained on scientific, biomedical and clinical text to identify sequences of biomedical entities. We find that the off-the-shelf model transfers poorly to our target data which stems from social media. We provide the evaluation results for this experiment in Appendix \ref{appendix-analysis-NER-results}. Therefore, we train a custom NER model in spaCy on
the \textsc{BEAR} dataset. We create an empty model using
spacy.blank() and pass the language ID ``en'' for English.  We provide
the train/test splits and configuration file we use to train the model which includes
all settings and hyperparameters here: \url{https://tinyurl.com/bear-ner}

\subsection{Main Claim Detection}
We fine-tune
RoBERTArg\footnote{\url{https://huggingface.co/chkla/ roberta- argument}}
to classify texts as \textsc{Claim} vs.\ \textsc{non-Claim} using the
Causal Claim data. We create a train-validation split of 85/15~\%. We
train for 5 epochs with a batch size of 16, 409 training steps per
epoch, 136 warmup steps and a weight decay of 0.01. We use the same
learning rate that was used in fine-tuning the underlying RoBERTArg
model, i.e., a learning rate of 2.3102e-06.  We evaluate the model
every 500 steps using the validation set. After training, we use the
model with the best performance on the validation set to make a
prediction for each claim candidate.

\subsection{Entity linking}
We use the \textit{EntityLinking}
component in ScispaCy \citep{neumann-et-al-2019} and \textit{en\_core\_sci\_sm} as the underlying model\footnote{\url{https://allenai.github.io/scispacy/}}. For each entity, the model maps the mention to the associated concept within UMLS \citep{bodenreider-2004}. We include the option to resolve abbreviations and leave the other configuration parameters at their default values.

\section{Analysis}
\label{appendix:analysis}
We provide an evaluation and analyses of individual pipeline components to better understand the capabilities of the modules.
\subsection{Evaluation Setup}
\label{appendix:analysis-eval-setup}
\paragraph{NER.}
Entity recognition consists of two subtasks: (a) identifying the span of an entity and (b) predicting the entity class.
Consequently, we evaluate the NER component of our pipeline in two modes. In the \textit{strict} mode, the entity span and the entity class have to be identical to the gold data. In the \textit{relaxed} mode, the entity span has to be identical to the gold data, entity class labels is ignored.

Note that the off-the-shelf ScispaCy \citep{neumann-et-al-2019} model that we compare against only labels the entity span and not the entity class. Therefore, we can only evaluate its performance in the \textit{relaxed } mode.

Further note that we need to map certain entity classes between the Co\textsc{Vert} and the \textsc{BEAR} dataset. To align Co\textsc{Vert}  with \textsc{BEAR}, we map \textit{Medical Condition} to \textit{med\_C}, \textit{Treatment} to \textit{treat\_therapy}, and \textit{OTHER} to other, respectively. The Co\textsc{Vert}  dataset further contains the class \textit{Symptom/Side-effect}, which corresponds to the class \textit{med\_C} of the \textsc{BEAR} dataset. Therefore, we map the class \textit{Symptom/Side-effect} to the class \textit{med\_C}. Entities which have been labeled in \textsc{Bear}, but not in Co\textsc{Vert}, are ignored for the evaluation.

We report the macro-average of precision, recall and \F for both modes.

\paragraph{Main claim detection.}
We evaluate the prediction of the model on the held-out test set from the \textsc{Causal claim} data.
We report precision, recall and \F for both classes (\textsc{Claim} vs. \textsc{non-claim}) the as well as the macro-average. 

\subsection{Results}
\subsubsection{NER}
\label{appendix-analysis-NER-results}
We evaluate the performance of the NER component within our pipeline. Table \ref{tab:NER-eval} reports the results for the strict and relaxed evaluation mode. First, we evaluate the performance on the unseen test split of the \textsc{Bear} data -- the dataset we use for training the model. To gauge how well it transfers to our target data, we evaluate the performance for the entity predictions in \textsc{C}o\textsc{Vert}. We compare the performance of our custom model to the performance of the pre-trained ScispaCy model. 

For the \textsc{Bear} data, our model reaches an average \F of 0.41
for the strict evaluation mode. Note that in this mode only exact span and entity type matches count as true positives. If we relax this condition and disregard the entity type, the model achieves an \F-score of 0.51.
When moving to a slightly different type of input text, i.e., the \textsc{C}o\textsc{Vert} data, the average \F-scores for the strict and relaxed evaluation modes reach 0.34 and 0.38, respectively. 

Compared to our custom model, the performance of the off-the-shelf model from ScispaCy is much lower. For the relaxed mode, we observe a $\Delta$ in \F of 0.21 and 0.12 for the \textsc{Bear} and \textsc{C}o\textsc{Vert} data, respectively. This showcases the necessity of a customized model for NER in this setting.

Overall, this evaluation of the entity recognition shows moderate performance. Importantly, the results also indicate that improving this component is likely to improve the overall fact-checking performance.

\begin{table}
\renewcommand{\arraystretch}{0.97}
  \small\centering
  \setlength{\tabcolsep}{5pt}
  \begin{tabular}{llrrrrrr}
    \toprule
    & & \multicolumn{6}{c}{target data} \\
    \cmidrule(lr){3-8}
    & & \multicolumn{3}{c}{\textsc{BEAR}}	& \multicolumn{3}{c}{\textsc{C}o\textsc{Vert}} \\  
    \cmidrule(lr){3-5} \cmidrule(lr){6-8}
    model & eval. mode &P & R & \F & P & R & \F\\
    \cmidrule(lr){1-1}\cmidrule(lr){2-2}\cmidrule(lr){3-3}\cmidrule(lr){4-4}\cmidrule(lr){5-5}\cmidrule(lr){6-6}\cmidrule(lr){7-7}\cmidrule(lr){8-8}

    \multirow{2}{*}{ScispaCy} & strict &	- &- & - & - & - & - \\   
    & relaxed & .2 & .61 & .3 & .16 & .72 & .26\\
    \cmidrule(lr){1-1} \cmidrule(lr){2-2} \cmidrule(lr){3-5} \cmidrule(lr){6-8}
    \multirow{2}{*}{Ours} & strict & .46 &.37 & .41 &   .26 & .51 & .34  \\
    & relaxed & .56 & .46 & .51 &    .29 & .57 & .38\\
    
    \bottomrule
  \end{tabular}
  \caption{Evaluation of our NER module for the test split of the \textsc{BEAR} dataset and the \textsc{C}o\textsc{Vert} data. We report the macro average precision (P), recall (R) and \F across all entity classes. We report results for a strict and a relaxed evaluation mode. We compare against the performance of an off-the-shelf ScispaCy \citep{neumann-et-al-2019} model (\textit{en\_core\_sci\_sm}). This model only labels the entity span, not the entity class. Therefore, we only evaluate in the relaxed mode.}
  \label{tab:NER-eval}
\end{table}

\subsubsection{Main claim detection}
We evaluate the performance of the claim detection model on the held-out test set. We report the results in Table \ref{tab:main-claim-model-performance}. We can see that the model successfully differentiates claims from non-claims (\F-scores of 0.94 and 0.99, respectively).

\begin{table}
\small\centering
    \begin{tabular}{lrrr}
    \toprule
	class & \multicolumn{1}{c}{P} & \multicolumn{1}{c}{R} & \multicolumn{1}{c}{\F}\\
	\cmidrule(lr){1-1}\cmidrule(lr){2-2}\cmidrule(lr){3-3}\cmidrule(lr){4-4}
	Non-claim & 0.98 & 0.99 & 0.99\\
	Claim & 0.95 & 0.93 &  0.94\\
	\cmidrule(lr){2-4}
	macro av. &0.97 & 0.96 & 0.96 \\

    \bottomrule
    \end{tabular}
    \caption{Performance (precision (P), recall (R), \F) of the claim detection model for \textsc{Causal claims} test set.}
    \label{tab:main-claim-model-performance}
\end{table}

\subsection{Analysis of \textit{healthver} prediction}
\label{sec-error-analysis-healthver}
We want to understand why the \textit{healthver} model behaves unexpectedly compared to the other models (refer to Table~\ref{tab:multivers-on-covert}). We saw that providing the automatically extracted claim leads to a slight performance decrease compared to inputting the full tweet, while the claims extracted using gold entities were more successfully checked. We hypothesize that for this model, the automatic extraction either removed relevant pieces of the input that it relied on previously for a successful prediction or it may have introduced irrelevant noise. Therefore, we compare the predictions of this model for our \textit{ner$+$core-claim} inputs to the claims built on gold-labeled entities $\textrm{condense}_{\textrm{seq}}$. Note that we compare the predictions which are not necessarily in line with the gold label.
\paragraph{Label distribution.} Table \ref{tab:label-distribution} reports the distribution of  predicted labels for both input types.
The \textsc{NEI} class increases substantially (115 to 158 predicted instances) while \textsc{Support} and \textsc{refute} become less frequent. This indicates that the claims become less checkable as \textsc{Nei} means a lack of information to support or refute the claim.
\paragraph{Label flips.} To better understand which instances cause the model to predict a different verdict, we present the number of label transitions between the predictions for the gold-labeled entity claims and the predictions for our pipeline claims (\textit{ner$+$core-claim}) in Table~\ref{tab:label-flips-examples}.
From those results we can observe that for a substantial amount of instances (161) the predicted label actually does not shift. For 90 instances, we observe a label shift.

Most notably, claims that were supported and refuted when inputting the gold-entity claims, get classified as \textsc{Nei} when we input our extracted claims (46 and 18, respectively). In an introspection of this transition type, we observe that in cases, the automatic pipeline failed to detect the main claim, potentially rendering the evidence useless. Refer to Claims~3 and 4 in Table~\ref{tab:label-flips-examples} for examples.

Flipped verdicts (\textsc{Supports} to \textsc{refutes} or vice versa) are less frequent. We observe a total of 11 instances. Refer to Claims~1 and 2 in Table~\ref{tab:label-flips-examples}.

We observe 15 cases in which the label flips to the correct gold label when we input our claim as opposed to the gold-entity-based claim. In the manual introspection, we observe many cases in which the claim from the pipeline slightly extends the context compared to the gold entity claim. Refer to Claims~5 and 6 in Table~\ref{tab:label-flips-examples} for two examples.

For cases with consistent labels, we find that many instances either are identical to the claim extracted using gold entities (see Table~\ref{tab:label-flips-examples}, Example~7a) or only small amounts of context is added (see Table~\ref{tab:label-flips-examples}, Ex.~8).

This being said, we also observe cases in which the gold-entity and our predicted claim do not overlap and yet, the verdict stays consistent (Ex.~7b). 
This emphasizes the need to further improve the main claim detection step and leads us to hypothesize that this module may be another reason for the limited performance of this model. It appears that the \textit{healthver} model is particularly sensitive to this component being somewhat unreliable and error propagation in general.

\begin{table}
\small\centering
    \begin{tabular}{lrrr}
    \toprule
    & \multicolumn{3}{c}{\# Predicted labels}\\
    \cmidrule(lr){2-4}
    input claim &\textsc{Support} & \textsc{Refute} & \textsc{NEI} \\
	\cmidrule(lr){1-1}\cmidrule(lr){2-2}\cmidrule(lr){3-3}\cmidrule(lr){4-4}
	gold entities & 110 & 39 & 115\\
	\textit{ner$+$core-claim} & 69 & 24 & 158 \\
	    
    \bottomrule
    \end{tabular}
    \caption{Labels predicted by the \textit{healthver} model for claims extracted using $\textrm{condense}_{\textrm{seq}}$ based on gold entities and our pipeline (\textit{ner$+$core-claim}). Note that there are 13 claims more in the gold-entity setting compared to \textit{ner$+$core-claim} inputs. These are cases for which the NER module predicted $\leq 1$ entity.}
    \label{tab:label-distribution}
\end{table}

\begin{table*}
\small\centering
    \begin{tabularx}{\textwidth}{llrXXl}
    \toprule
    & & & \multicolumn{3}{c}{example}\\
    \cmidrule(lr){4-6}
    id & transition & \# inst. & gold-ent-claim & ner$+$core-claim & gold\\
    \cmidrule(lr){1-1}\cmidrule(lr){2-2}\cmidrule(lr){3-3}\cmidrule(lr){4-4} \cmidrule(lr){5-5} \cmidrule(lr){6-6}
    1 & S-R & 7 &Oral contraceptives cause more blood clots & blood clots and nobody is doing anything about that!!! Like 1 per 1,000 compared to basically 1 per MILLION with the Covid vaccine & S\\
    2 & R-S &4 & COVID-19 vaccines can cause side effects & Vaccine reactions are rare. Info about side effects & S\\
    3 & S-NEI & 46& COVID-19 1) directly causes viral pneumonia & pneumonia 3) can result in intubation & S\\
    4 &R-NEI & 18& 5G causes covid & vaccines cause infertility \& autism & R\\
    5 & NEI-S & 12 & live virus that causes covid-19 & vaccines don't use the live virus that causes covid-19 & S \\
    6 & NEI-R & 3 & masks cause plague & masks cause plague... fauci knows... masks promote bacteria... and not the good kind... sinus & R \\
	\cmidrule(lr){2-6}
	7a & S-S & \multirow{2}{*}{53} & covid vaccine doesn't cause fertility issues & covid vaccine doesn't cause fertility issues & S \\
	7b & S-S & & all brands of the vaccine can cause problems & death rate of COVID is said to be 10\%. It is probable that some vaccines & S \\
	8 & R-R  & 14 & Wearing a mask does cause disease & Wearing a mask does cause disease, harm the immune system & R\\  
	9 & NEI-NEI &94 & Auto-Immune disease causes the white blood cells that normally protect your body from invaders to turn around and attack your cells, tissues and organs & Auto-Immune disease causes the white blood cells that normally protect your body & S \\
    \bottomrule
    \end{tabularx}
    \caption{Label transitions as predicted by the \textit{healthver} model for claims extracted using $\textrm{condense}_{\textrm{seq}}$ based on gold entities (gold-ent-claim) and our pipeline (\textit{ner$+$core-claim}). We provide example instances for each type of label transitions along with the gold label for the fact-checking verdict. }
    \label{tab:label-flips-examples}
\end{table*}

\subsection{Entity Linking}
\label{sec:analysis-linking}

\paragraph{Number of established mappings.}
There are no gold annotated mappings for the medical entities in the
Co\textsc{Vert} dataset that would allow for a full evaluation. We therefore
approximate one aspect of the quality of the entity linking module by analyzing the
number of entities that are being linked to any concept in the first
place.
Out of 719 entity mentions the linking module established mappings for 495 instances (68.8 \%). We provide insights from an error analysis in the following section.

\paragraph{Error analysis.}
We aim to understand the type of error patterns introduced by the entity linking module. We analyze predicted links for a randomly drawn sample of 25 entities. We manually categorize the predicted concepts with regard to four properties. Table \ref{tab:linking-error-counts} reports the results as well as examples. \textit{correctly linked} instances are mapped to the appropriate concept within UMLS. \textit{Incorrect but related link} include instances which are mapped incorrectly, but the concept is related. \textit{incorrect and unrelated link} include cases in which the linking is incorrect and also unrelated.

The analysis shows that the majority of mentions are linked to the correct (15 out of 25 instances) or at least a related (6 out of 25 instances) UMLS concept. Four instances within our sample were mapped to an unrelated UMLS concept. 

While the majority of cases within our sample are normalized correctly, this module potentially introduces many errors. Note that as pointed out before about 30 \% of entities are not linked and consequently not replaced at all. In addition, an incorrectly mapped and replaced mention, even if the concept might be closely related, may change the meaning of a claim drastically. Take the following example claim: `COVID cause of breathlessness'. While \textit{breathlessness} is correctly mapped to \textit{dyspnea}, \textit{COVID} is linked to and subsequently replaced by an unrelated concept: `Covi Anxiety Scale Clinical Classification cause of dyspnea'. This leads us to believe that the unreliability of the linking module is the main reason why the verdict prediction performance for the normalized claims is comparably low.
 
\begin{table}
\renewcommand{\arraystretch}{0.9}
\small
    \setlength{\tabcolsep}{4pt} 
    \begin{tabularx}{\columnwidth}{XrXX}
    \toprule
	error type & \# & mention & pred. concept \\
	\cmidrule(lr){1-1}\cmidrule(lr){2-2}\cmidrule(lr){3-3}\cmidrule(lr){4-4}
	\textit{correctly linked} & 15 & glandular fever & Infectious Mononucleosis\\
	\textit{incorr., related} & 6 & fibro flare & Fibromyalgia\\
	\textit{incorr. \& unrelated} & 4 & COVID & Covi Anxiety Scale [...]\\

    \bottomrule
    \end{tabularx}
    \caption{Number of error types within a sample of 25 entities along with examples.}
    \label{tab:linking-error-counts}
\end{table}

\end{document}